\documentclass{article}
\usepackage{spconf,amsmath,epsfig}

\usepackage{algorithm}
\usepackage{algorithmicx}
\usepackage{amsmath}
\usepackage{amsfonts}
\usepackage{graphics}
\usepackage{algpseudocode}
\usepackage{booktabs}

\usepackage{multirow}
\usepackage{subfigure}
\usepackage{epsfig}
\usepackage{enumitem}
\usepackage{mathrsfs}
\usepackage{bm}
\usepackage{graphicx}
\usepackage{epstopdf}
\usepackage{verbatim}
\usepackage{array}

\usepackage[switch]{lineno}

\let\OLDthebibliography\thebibliography
\renewcommand\thebibliography[1]{
  \OLDthebibliography{#1}
  \setlength{\parskip}{0pt}
  \setlength{\itemsep}{0pt plus 0.3ex}
}

\begin{document}\sloppy

\def\x{{\mathbf x}}
\def\L{{\cal L}}

\title{Weakly-Supervised Online Hashing}
%
\name{Yu-Wei Zhan, Xin Luo$^{\ast}$, Yu Sun, Yongxin Wang, Zhen-Duo Chen, Xin-Shun Xu}
\address{School of Software, Shandong University, Jinan, China \thanks{$^{\ast}$ Corresponding author. Email: luoxin.lxin@gmail.com. 978-1-6654-3864-3/21/\$31.00 ©2021 IEEE}}

\maketitle

\begin{abstract}
With the rapid development of social websites, recent years have witnessed an explosive growth of social images with user-provided tags which continuously arrive in a streaming fashion. Due to the fast query speed and low storage cost, hashing-based methods for image search have attracted increasing attention. However, existing hashing methods for social image retrieval are based on batch mode which may violate the nature of social images, i.e., social images are usually generated periodically or collected in a stream fashion. Although there exist many online hashing methods, they either adopt unsupervised learning which ignore the relevant tags, or are designed in the supervised manner which needs high-quality labels. In this paper, to overcome the above limitations, we propose a new method named Weakly-supervised Online Hashing (WOH). In order to learn high-quality hash codes, WOH exploits the weak supervision, i.e., tags, by considering the semantics of tags and removing the noise. Besides, we develop a discrete online optimization algorithm for WOH, which is efficient and scalable. Extensive experiments conducted on two real-world datasets demonstrate the superiority of WOH.
\end{abstract}

\section{Introduction}
\label{sec:intro}
Recently, there has been an explosive growth of data on the Internet. As most traditional similarity search methods are not applicable to large-scale data, hashing-based Approximate Nearest Neighbor techniques have been proposed and widely studied \cite{weiss2009spectral,shen2015supervised,kang2016column,cui2020efficient}. These methods learn hash functions that can transform high-dimensional data into short strings of binary codes while preserving the similarity of the original data. As a result, the storage cost can be reduced and the retrieval can be performed efficiently in the Hamming space.

Most existing hashing methods are batch-based, which means that they need to accumulate all data and retrain the hash functions when new data comes. However, in real-world Internet, data usually become available continuously as streams \cite{leng2015online,lin2019towards,lin2020hadamard}, making those batch-based methods inefficient. Recently, to overcome the limitations, several hashing methods are designed in an online manner. Roughly speaking, existing online hashing can be divided into supervised methods \cite{lin2019towards,lin2020hadamard,cakir2017online} and unsupervised ones \cite{leng2015online,chen2017frosh}. However, none of existing online hashing methods is specially designed for weakly-supervised images, i.e., social images with tags.

As social images with user-provided tags are being generated continuously and prevalent on the Web, these weakly-supervised images naturally come in a streaming fashion. Thus, it is essential to design approaches which can fulfill the need for online social image retrieval. However, compared to full supervision, i.e., the clean class labels, user-provided tags are weak and imperfect. Hence, directly using the tags as the supervised information and imputing them to online supervised hashing methods may lead to suboptimal performance. Unsupervised online hashing could support retrieving social images without considering tags. However, these freely available tags naturally contain relevant semantics of images and leaving tags out of consideration may lead to information loss and poor performance. Although there exist some weakly-supervised hashing methods (also known as social image hashing) \cite{cui2020efficient,gattupalli2019weakly,li2020weakly}, which are specially designed to learn hash codes with the help of weakly-supervised tagging information, all of them are batch-based and not able to support the online setting.

To tackle the issues mentioned above, we propose a novel hashing method named Weakly-supervised Online Hashing (WOH), which incorporates hash code learning, the online learning, and the weakly-supervised information mining into one unified framework. To excavate the semantics in the user-provided tags, construct the connections between tags and hash codes and simultaneously avoid the adverse impacts from imperfect tags by introducing the $\ell_{2,1}$-norm. Moreover, an efficient discrete online optimization is proposed. The main contributions of WOH are summarized as follows:

1) By considering the denoised tags, image level semantic representations, and the visual features of images, WOH can generate the hash codes of social images in online scenarios that preserve the semantic information from tags and eliminate the negative effects of noisy tags.

2) We propose an iterative online optimization algorithm. Its time complexity is linear to the newly arriving data size per round, making WOH efficient and scalable. Besides, during optimization, hash codes can be discretely learnt with the binary constraints maintained.

3) Extensive experiments are conducted over two widely-used benchmark datasets. The results demonstrate the superiority of WOH over several state-of-the-art online hashing methods and social image hashing methods.

4) As far as we know, weakly-supervised online hashing is a completely new domain and WOH may be the first attempt to study this novel and valuable topic. The code is released (https://github.com/smile555/WOH).

\section{Our Method}
\subsection{Notations}
Suppose the training data comes at a streaming manner. At the $t$-th round, a new data chunk of social images $\vec{\textbf{X}}^{(t)}\in\mathbb{R}^{n_t\times d}$ is added to the training set, where $n_t$ is the size of new data chunk and $d$ is the dimensionality of image feature. Please note that, in social image hashing, no labels are provided and the tags $\vec{\textbf{Y}}^{(t)}\in \{0, 1\} ^{{n_t} \times {c}}$ are viewed as weak supervision information for learning. Specifically, $c$ is the number of tags, $\vec{\textbf{Y}}^{(t)}_{ij}=1$ if the $i$-th image at the $t$-th round is associated with tag $j$ and 0 otherwise. Correspondingly, the already accumulated old data, which has been accumulated before round $t$, is represented as $\tilde{\textbf{X}}^{(t)}\in\mathbb{R}^{N_{t-1}\times d}$ and the user-provided tags is denoted as  $\tilde{\textbf{Y}}^{(t)}$, where $N_{t-1}=\sum_{i=1}^{t-1}n_i $ is the size of the existing data. The goal of weakly-supervised online hashing is to learn $r$-bit binary hash codes $[\tilde{\textbf{B}}^{(t)};\vec{\textbf{B}}^{(t)}] \in {\{-1, 1\} ^{{N_t} \times {r}}}$, where $\tilde{\textbf{B}}^{(t)}$ is the hash codes of the existing old data and $\vec{\textbf{B}}^{(t)}$ is the hash codes of the newly coming data, respectively.

\subsection{Model Formulation}
\textbf{Weak Supervision (tags) Processing.} The idea that regressing hash codes to supervision information \cite{shen2015supervised} or regressing supervision information to hash codes \cite{gui2017fast} has been widely accepted and proven to be effective in hashing literature. Thus, the corresponding objective at round $t$ can be given:
\begin{equation}\label{eq1}
\begin{split}
& \mathcal{O}_{reg}\! =\!  \parallel \tilde{\textbf{Y}}^{(t)}\! - \! \tilde{\textbf{B}}^{(t)}\textbf{W}^{(t)} \parallel_{2,1}\!   +\!
\parallel \!\vec{\textbf{Y}}^{(t)}\! -\!  \vec{\textbf{B}}^{(t)}\!\textbf{W}^{(t)}\! \parallel_{2,1}, \\
\end{split}
\end{equation}
where $\textbf{W}^{(t)}$ is a projection from hash codes to weak supervision. As weakly-supervised tagging information may contain noise, directly building up relationship among hash codes and tags tend to be suboptimal. Thus, in Eq.(\ref{eq1}), we adopt the $\ell_{2,1}$-norm \cite{nie2010efficient}, which has demonstrated to be effective to alleviate the noise problem.

In order to effectively mine the user-provided tags, we try to represent the image with the high-level tag semantics. Through taking advantages of the superior ability endowed by the NLP techniques \cite{mikolov2013efficient}, we can project tags into a word embedding space. In this way, each tag word is represented as a vector embedding. Thereafter, we further aggregate the vector embeddings of one image by average pooling and get the \emph{image level semantic representation} $\textbf{z}_i$ for image $i$. As have been analysed in previous works in hashing literature \cite{gattupalli2019weakly,guan2018tag}, the advantages of using word2vec tool can be two-sided: 1) the zero-shot problem caused by tag incompleteness could be alleviated; 2) the noisy tags could be suppressed to some extent. However, the embedding model is off-the-shelf and may bring in potential semantic shift between the tags in our task and words in its training corpus. To consider this problem,  we further refer to the visual information for help. Specifically, at round $t$, the objective function to learn hash codes from the high-level tag semantics along with the visual features can be formulated as follows,
\begin{equation}\label{eq2}
\begin{split}
& \mathcal{O}_{sem} = \beta\!\parallel\!\tilde{\textbf{X}}^{(t)}\! -\! \tilde{\textbf{B}}^{(t)}\!\textbf{U}^{(t)} \!\parallel^2_F \! +\! \beta\! \parallel\! \vec{\textbf{X}}^{(t)}\! -\! \vec{\textbf{B}}^{(t)}\!\textbf{U}^{(t)}\! \parallel^2_F\\
& +\!\theta\!\parallel\! \tilde{\textbf{Z}}^{(t)}\!-\!\tilde{\textbf{B}}^{(t)}\!\textbf{V}^{(t)}\!  \parallel^2_F  \!+\! \theta \!\parallel\! \vec{\textbf{Z}}^{(t)}\! -\! \vec{\textbf{B}}^{(t)}\!\textbf{V}^{(t)}\! \parallel^2_F,\\
\end{split}
\end{equation}
where $\beta$ and $\theta$ are trade-off parameters, $\tilde{\textbf{Z}}^{(t)}$ and $\vec{\textbf{Z}}^{(t)}$ are the image level semantic  representations of old data and new data, $\textbf{U}^{(t)}$ and $\textbf{V}^{(t)}$ are both auxiliary variables, and $\parallel \cdot \parallel_F$ is the Frobenius norm. The first two terms and the last two terms embed the visual information and the high-level tag semantics, respectively.

\textbf{Hash Function Learning.} The hash function is learnt to transform the out-of-sample images into hash codes. For example, at the $t$-th round, when a new query sample comes, we need to generate its hash code by $\textbf{B}^{(t)}_q=sign(\textbf{X}_q\textbf{P}^{(t)})$, where $\textbf{P}^{(t)}$ is the projection matrix of hash function. For this purpose, we define the hash function learning loss at round $t$:
\begin{equation}\label{eq3}
\begin{split}
\mathcal{O}_{fun}\! =\! \mu \parallel \! \tilde{\textbf{B}}^{(t)}\! -\! \tilde{\textbf{X}}^{(t)}\textbf{P}^{(t)} \! \parallel^2_F \! +\!  \mu \parallel\! \vec{\textbf{B}}^{(t)}\! -\! \vec{\textbf{X}}^{(t)}\textbf{P}^{(t)}\!  \parallel^2_F,
\end{split}
\end{equation}
where $\mu$ is a parameter. In online hashing settings, relying only on the newly arrived data to update hash functions may lose the information of existing data and become suboptimal. Thus, we consider both the old accumulated data (the first term) and the newly arrived data (the second term) in Eq.(\ref{eq3}).

\textbf{Overall Objective Function.} Combining Eq.(\ref{eq1}), Eq.(\ref{eq2}), and Eq.(\ref{eq3}), the overall objective function can be written as,
\begin{equation}\label{overall}
\begin{split}
&\min_{\vec{\textbf{B}}^{(t)},\textbf{W}^{(t)},\textbf{U}^{(t)},\textbf{V}^{(t)},\textbf{P}^{(t)}} \mathcal{O}_{reg}\! +\! \mathcal{O}_{sem}\! +\! \mathcal{O}_{fun}\! \\
& +\! \alpha R( \textbf{W}^{(t)} ,\textbf{U}^{(t)},\textbf{V}^{(t)},\textbf{P}^{(t)}),  s.t.\;\vec{\mathbf{B}}^{(t)} \in {\{-1,1\} ^{{n_t} \times r }}\\
\end{split}
\end{equation}
where $\alpha$ is a parameter, $R(\cdot)=\parallel \cdot \parallel^2_F$ is the regularization term to avoid overfitting, and $\parallel \cdot \parallel_F$ is the Frobenius norm.

Besides, to capture the nonlinear characteristics, the kernel features $\phi {(\textbf{X})}$ are adopted to replace the original image features. Specifically, the RBF kernel mapping is adopted, i.e., $\phi {(\textbf{x})}= exp(\frac{-\parallel {\textbf{x}}-\textbf{a}_i \parallel^2_2}{2\sigma^2})$, where ${\{\textbf{a}_i\}}_{i=1}^m$ denotes the randomly selected $m$ anchor points from the training samples at the first round and $\sigma$ denotes the kernel width calculated by $\sigma = \frac{1}{nm}\sum_{i=1}^{n}\sum_{j=1}^{m}\parallel {\textbf{x}_i}-\textbf{a}_j \parallel_2$.


\subsection{Efficient Discrete Online Optimization}
To solve the optimization problem in Eq. (\ref{overall}), we propose a five-step iterative scheme as follows.

\textbf{$\textbf{U}^{(t)}$ Step.} We fix $\textbf{W}^{(t)}$, $\textbf{V}^{(t)}$, $\textbf{P}^{(t)}$, and $\vec{\textbf{B}}^{(t)}$, and update $\textbf{U}^{(t)}$ by solving the following objective function,
\begin{equation}\label{U_1}
\begin{split}
&\min_{\textbf{U}^{(t)}}\beta\parallel\phi{(\tilde{\textbf{X}}^{(t)})}-\tilde{\textbf{B}}^{(t)}\textbf{U}^{(t)} \parallel^2_F \\
&+\beta \parallel\phi{(\vec{\textbf{X}}^{(t)})}-\vec{\textbf{B}}^{(t)}\textbf{U}^{(t)}\parallel^2_F+ \alpha \parallel \textbf{U}^{(t)}\parallel^2_F.
\end{split}
\end{equation}
By setting the derivative of Eq.(\ref{U_1}) w.r.t. $\textbf{U}^{(t)}$ to zero, we have,
\begin{equation}\label{U_2}
\textbf{U}^{(t)}=(\textbf{C}_1^{(t)}+\frac{\alpha}{\beta}\textbf{I})^{-1}\textbf{C}_2^{(t)},
\end{equation}
where $\textbf{C}_1^{(t)} = \tilde{\textbf{B}}^{(t)\top}\tilde{\textbf{B}}^{(t)} + \vec{\textbf{B}}^{(t)\top}\vec{\textbf{B}}^{(t)}$ and $\textbf{C}_2^{(t)} = \tilde{\textbf{B}}^{(t)\top}\phi{(\tilde{\textbf{X}}^{(t)})} + \vec{\textbf{B}}^{(t)\top}\phi{(\vec{\textbf{X}}^{(t)})}.$ Apparently, $\textbf{C}_1^{(t)}$ can be transformed as follows,
\begin{equation}\label{U_4}
\begin{split}
\textbf{C}_1^{(t)}\! &=\! \tilde{\textbf{B}}^{(t)\top}\!\tilde{\textbf{B}}^{(t)}\! + \! \vec{\textbf{B}}^{(t)\top}\!\vec{\textbf{B}}^{(t)}\\
&\!=\!\left[\!\tilde{\textbf{B}}^{(t-1)}\! \; ; \;\!\vec{\textbf{B}}^{(t-1)}\!\right]^{\top} \!\left[\!\tilde{\textbf{B}}^{(t-1)}\! \; ; \;\!\vec{\textbf{B}}^{(t-1)}\!\right] + \!\vec{\textbf{B}}^{(t)\top}\!\vec{\textbf{B}}^{(t)}\\
&\!=\!\textbf{C}_1^{(t-1)}\! +\! \vec{\textbf{B}}^{(t)\top}\!\vec{\textbf{B}}^{(t)}.
\end{split}
\end{equation}
Similarly, we have $\textbf{C}_2^{(t)} = \textbf{C}_2^{(t-1)} + \vec{\textbf{B}}^{(t)\top}\phi{(\vec{\textbf{X}}^{(t)})}$.

\textbf{$\textbf{P}^{(t)}$ Step and $\textbf{V}^{(t)}$ Step.} As the optimizations of these two variables are very similar, we discuss them together. By setting the partial derivative of Eq.(\ref{overall}) w.r.t. $\textbf{P}^{(t)}$ to zero, we can update it by,
\begin{equation}\label{P_2}
\textbf{P}^{(t)}=(\textbf{C}_3^{(t)}+\frac{\alpha}{\mu}\textbf{I})^{-1}\textbf{C}_4^{(t)},
\end{equation}
where $\textbf{C}_3^{(t)} = \textbf{C}_3^{(t-1)} + \phi{(\vec{\textbf{X}}^{(t)})}^\top\phi{(\vec{\textbf{X}}}^{(t)})$ and $\textbf{C}_4^{(t)} = \textbf{C}_4^{(t-1)} + \phi{(\vec{\textbf{X}}^{(t)})}^\top\vec{\textbf{B}}^{(t)}$.

Similarly, the solution of $\textbf{V}^{(t)}$ is given,
\begin{equation}\label{V_2}
\textbf{V}^{(t)}=(\textbf{C}_1^{(t)}+\frac{\alpha}{\theta}\textbf{I})^{-1}\textbf{C}_5^{(t)},
\end{equation}
where $\textbf{C}_5^{(t)} = \textbf{C}_5^{(t-1)} + \vec{\textbf{B}}^{(t)\top}\vec{\textbf{Z}}^{(t)}$.

\textbf{$\textbf{W}^{(t)}$ Step.} With $\textbf{U}^{(t)}$, $\textbf{V}^{(t)}$, $\textbf{P}^{(t)}$, and $\vec{\textbf{B}}^{(t)}$ fixed, the problem of optimizing $\textbf{W}^{(t)}$ can be formulated as,
\begin{equation}\label{W}
\begin{split}
&\min_{\textbf{W}^{(t)}} \parallel \tilde{\textbf{Y}}^{(t)}- \tilde{\textbf{B}}^{(t)}\textbf{W}^{(t)} \parallel_{2,1} \\
& + \parallel \vec{\textbf{Y}}^{(t)}- \vec{\textbf{B}}^{(t)}\textbf{W}^{(t)} \parallel_{2,1} + \alpha \parallel \textbf{W}^{(t)}\parallel^2_F.
\end{split}
\end{equation}
To optimize the $\ell_{2,1}$-norm, the first two terms in Eq.(\ref{W}) are reformulated as,
\begin{equation}\label{W2}
\begin{split}
& Tr\big((\tilde{\textbf{Y}}^{(t)}- \tilde{\textbf{B}}^{(t)}\textbf{W}^{(t)})^\top\textbf{E}^{(t)}(\tilde{\textbf{Y}}^{(t)}- \tilde{\textbf{B}}^{(t)}\textbf{W}^{(t)}\big), \\
\end{split}
\end{equation}
\begin{equation}\label{W3}
\begin{split}
&Tr\big((\vec{\textbf{Y}}^{(t)}-\vec{\textbf{B}}^{(t)}\textbf{W}^{(t)})^\top\textbf{K}^{(t)} (\vec{\textbf{Y}}^{(t)}- \vec{\textbf{B}}^{(t)}\textbf{W}^{(t)}\big),
\end{split}
\end{equation}
where $\textbf{E}^{(t)}$ and $\textbf{K}^{(t)}$ are diagonal matrices, $\textbf{E}^{(t)}_{ii}=1/\parallel (\tilde{\textbf{Y}}^{(t)}- \tilde{\textbf{B}}^{(t)}\textbf{W}^{(t)})^i\parallel_2$, $\textbf{K}^{(t)}_{ii}=1/\parallel ( \vec{\textbf{Y}}^{(t)}- \vec{\textbf{B}}^{(t)}\textbf{W}^{(t)})^i\parallel_2$, and $(\cdot)^i$ denotes the $i$-th row of a matrix. Then, by setting the derivative w.r.t. $\textbf{W}^{(t)}$ to zero, we have,
\begin{equation}\label{W4}
\begin{split}
&\textbf{W}^{(t)}=({\textbf{D}_1}^{(t)}+\vec{\textbf{B}}^{(t)\top} \textbf{K}^{(t)}\vec{\textbf{B}}^{(t)}+\alpha \textbf{I})^{-1}\\
&\cdot({\textbf{D}_2}^{(t)}+\vec{\textbf{B}}^{(t)\top} \textbf{K}^{(t)}\vec{\textbf{Y}}^{(t)}),
\end{split}
\end{equation}
where ${\textbf{D}_1}^{(t)}=\tilde{\textbf{B}}^{(t)\top}\textbf{E}^{(t)}\tilde{\textbf{B}}^{(t)}$, ${\textbf{D}_2}^{(t)}=\tilde{\textbf{B}}^{(t)\top}\textbf{E}^{(t)}\tilde{\textbf{Y}}^{(t)}$, and $\textbf{I}$ is the identity matrix.

It is worth noting that, given the new coming data and the accumulated data, both ${\textbf{D}_1}^{(t)}$ and ${\textbf{D}_2}^{(t)}$ can be computed efficiently based the rules of block matrices. For example,
\begin{equation}\label{W5}
\begin{split}
&{\textbf{D}_1}^{(t)}\! =\! \tilde{\textbf{B}}^{(t)\top}\!\textbf{E}^{(t)}\!\tilde{\textbf{B}}^{(t)}\\
&\!=\!\left[\!\tilde{\textbf{B}}^{(t-1)}\!\; ; \;\!\vec{\textbf{B}}^{(t-1)}\!\right]\!
{\begin{bmatrix}
\textbf{E}^{(t-1)}&\textbf{0}\\
\textbf{0}&{\textbf{K}}^{(t-1)}
\end{bmatrix}}\!
\left[\!\tilde{\textbf{B}}^{(t-1)}\!\; ; \;\!\vec{\textbf{B}}^{(t-1)}\!\right]\!\\
&\!=\!{\textbf{D}_1}^{(t-1)}\! +\! \vec{\textbf{B}}^{(t-1)\top}\!\textbf{K}^{(t-1)}\!\vec{\textbf{B}}^{(t-1)}\!.
\end{split}
\end{equation}
Similarly, ${\textbf{D}_2}^{(t)} = {\textbf{D}_2}^{(t-1)} + \vec{\textbf{B}}^{(t-1)\top}\textbf{K}^{(t-1)}\vec{\textbf{Y}}^{(t-1)}$.

\begin{table*}[t] \scriptsize  \center  
\caption{The MAP results of various methods on MIRFlickr and NUS-WIDE at the last round. }
\setlength{\tabcolsep}{3.5mm}{
\label{map-table}\begin{tabular}{l ccccc ccccc}\toprule[1pt]
\multirow{2}{*}{Method}&\multicolumn{5}{c}{MIRFlickr}&\multicolumn{5}{c}{NUS-WIDE} \\\cmidrule(lr){2-6}
\cmidrule(lr){7-11} & 8 bits &16 bits&32 bits&64 bits&96 bits & 8 bits &16 bits&32 bits&64 bits&96 bits\\
\hline
 {SH \cite{weiss2009spectral}}&{0.6117}&{0.5994}&{0.5954}&{0.5990}&{0.5978}&{0.4221}&{0.4035}&{0.3819}&{0.3893}&{0.3802}\\
 {SDH \cite{shen2015supervised}}&{0.5940}&{0.6054}&{0.6307}&{0.6330}&{0.6353}&{0.4405}&{0.4910}&{0.4935}&{0.5057}&{0.5124}\\
 {COSDISH \cite{kang2016column}}&{0.5774}&{0.5786}&{0.5930}&{0.6085}&{0.6156}&{0.3828}&{0.4220}&{0.4711}&{0.5058}&{0.5317}\\
 {WDH \cite{cui2020efficient}}&{0.5800}&{0.6102}&{0.6265}&{0.6462}&{0.6565}&{0.4927}&{0.5168}&{0.5587}&{0.5856}&{0.6210}\\
 {OSH \cite{leng2015online}}&{0.6389}&{0.6418}&{0.6429}&{0.6474}&{0.6575}&{0.4687}&{0.4709}&{0.4749}&{0.4906}&{0.4998}\\
 {BOSDH \cite{lin2019towards}}&{0.5705}&{0.5676}&{0.5621}&{0.5626}&{0.5607}&{0.3214}&{0.3937}&{0.4014}&{0.4064}&{0.4090}\\
 {HMOH \cite{lin2020hadamard}}&{0.5788}&{0.5797}&{0.5839}&{0.5966}&{0.5984}&{0.3256}&{0.3309}&{0.3405}&{0.3409}&{0.3454}\\
 {WOH}&\textbf{0.6714}&\textbf{0.6833}&\textbf{0.6945}&\textbf{0.6967}&\textbf{0.6964}&\textbf{0.5521}&\textbf{0.5888}&\textbf{0.6043}&\textbf{0.6182}&\textbf{0.6221}\\
 \hline
\toprule[1pt]\end{tabular}} \label{map}\end{table*}

\begin{figure*}
\begin{minipage}{0.245\linewidth}\centering
\centerline{\includegraphics[height=3.8cm]{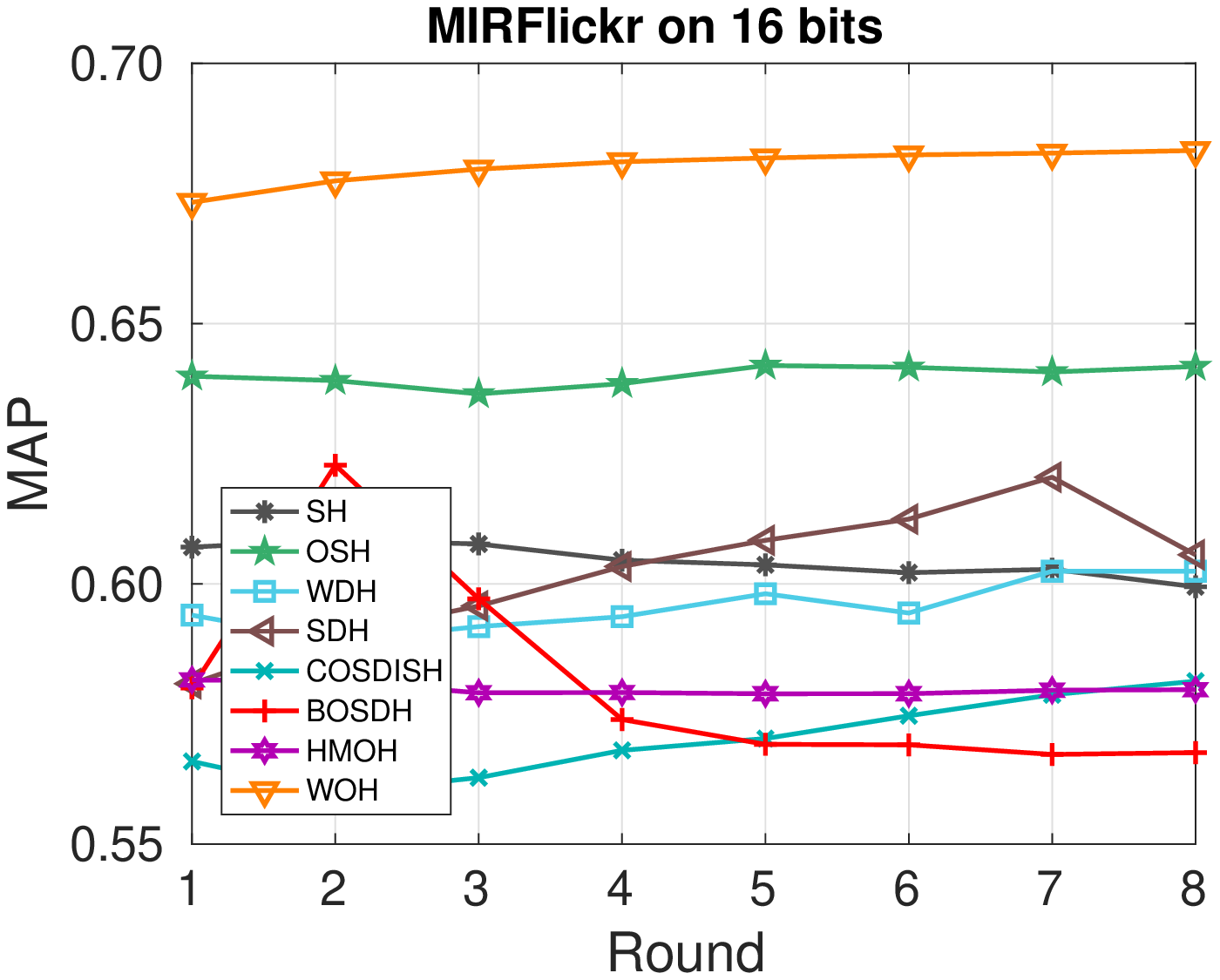}}
\end{minipage}
\begin{minipage}{0.245\linewidth}\centering
\centerline{\includegraphics[height=3.8cm]{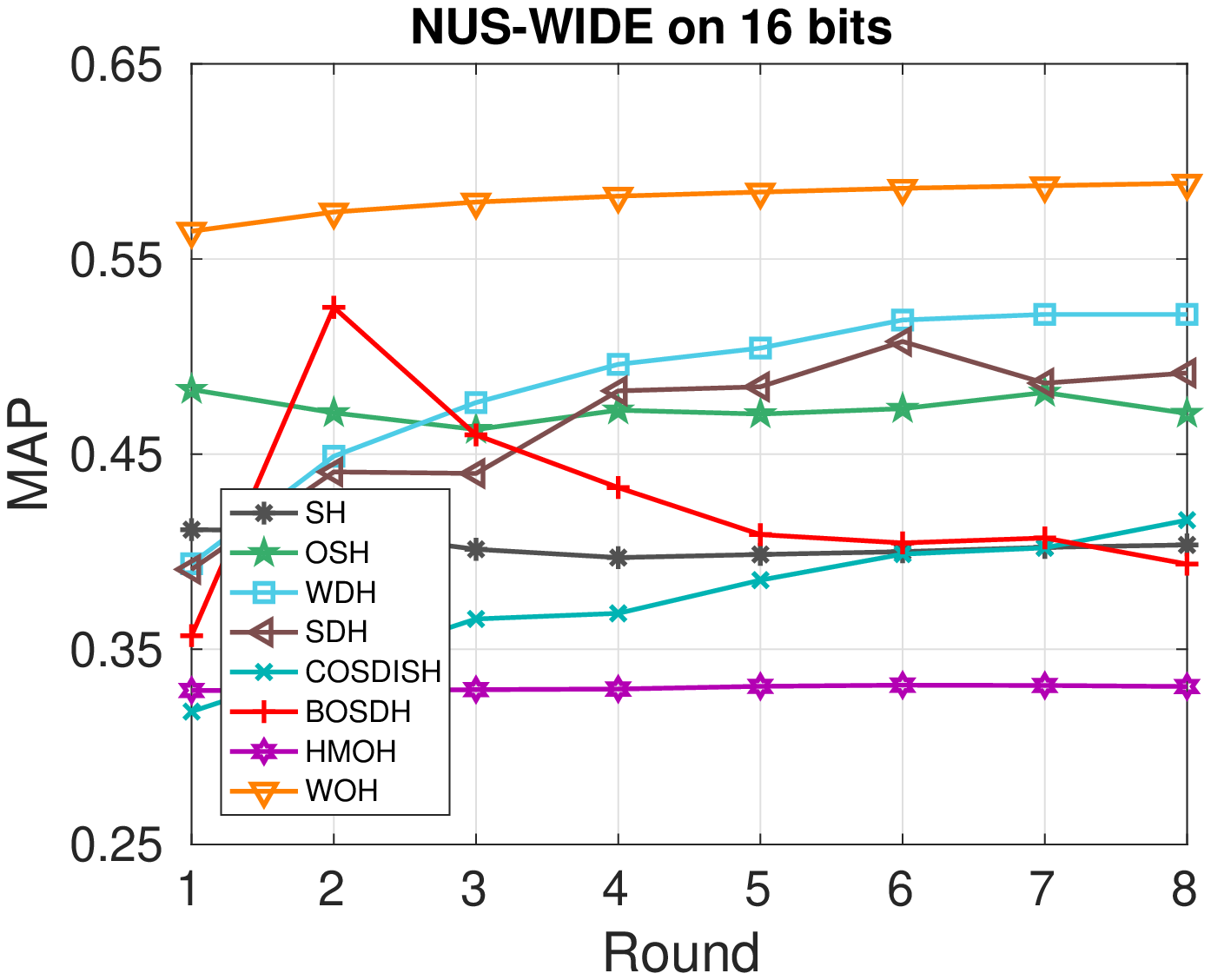}}
\end{minipage}
\begin{minipage}{0.245\linewidth}\centering
\centerline{\includegraphics[height=3.8cm]{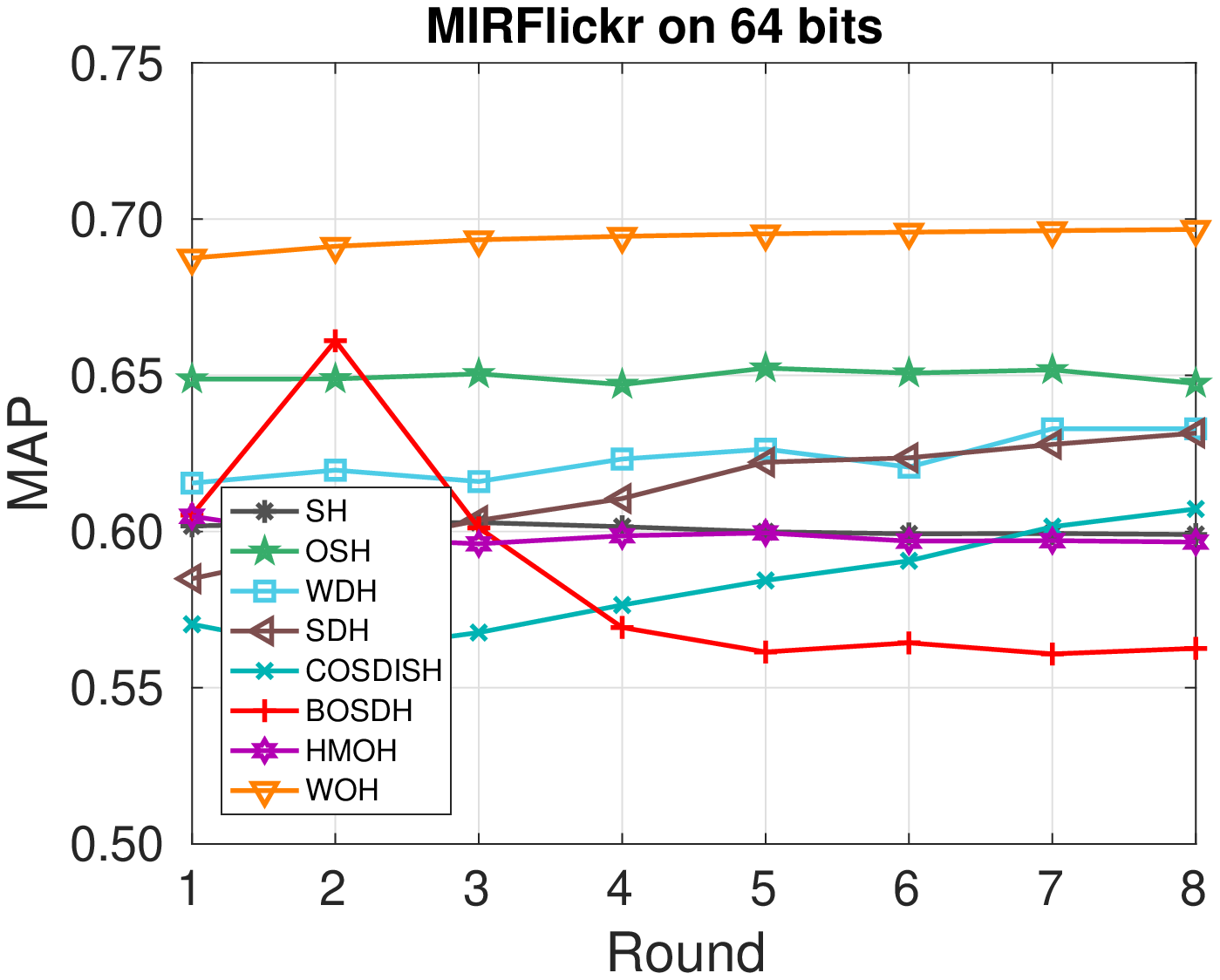}}
\end{minipage}
\begin{minipage}{0.245\linewidth}\centering
\centerline{\includegraphics[height=3.8cm]{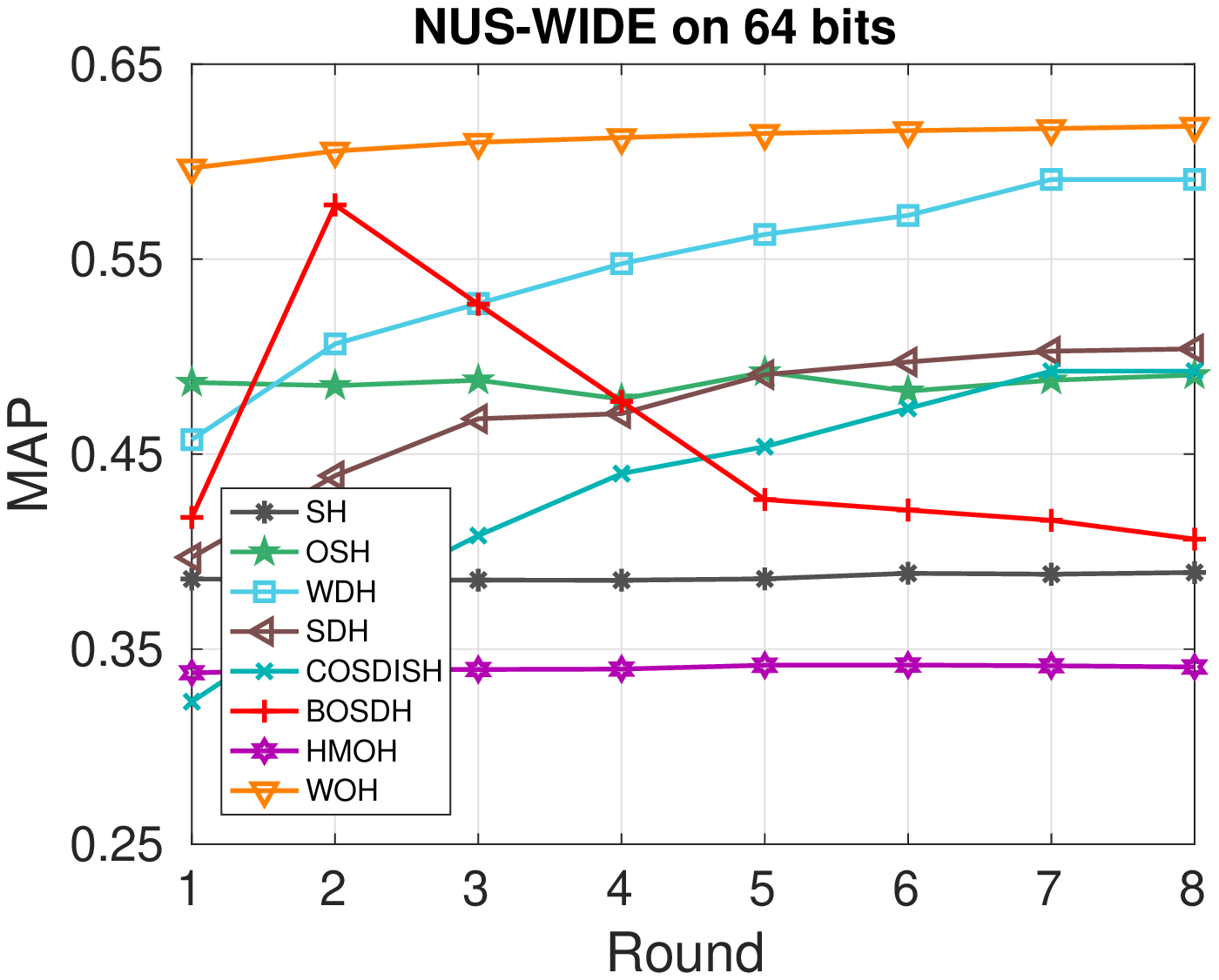}}
\end{minipage}\vspace{-0.3cm}
\caption{The MAP-round curves of various methods on MIRFlickr and NUS-WIDE.}\label{MAP-round}
\end{figure*}

\textbf{$\vec{\textbf{B}}^{(t)}$ Step.} Fixing other variables and rewriting Eq.(\ref{overall}), we can get the learning problem of $\vec{\textbf{B}}^{(t)}$,
\begin{equation}\label{B_2}
\begin{split}
&\min_{\vec{\textbf{B}}^{(t)}} \beta\parallel \vec{\textbf{B}}^{(t)}\textbf{U}^{(t)} \parallel_F^2 + \theta\parallel \vec{\textbf{B}}^{(t)}\textbf{V}^{(t)}\parallel_F^2 \\
&  + Tr(\textbf{W}^{(t)\top}\vec{\textbf{B}}^{(t)\top}\textbf{K}^{(t)}\vec{\textbf{B}}^{(t)}\textbf{W}^{(t)}) -2Tr(\vec{\textbf{B}}^{(t)\top}\textbf{Q}^{(t)}) \\
\end{split}
\end{equation}
where  $\textbf{Q}^{(t)}$ represents $\textbf{K}^{(t)}\vec{\textbf{Y}}^{(t)}\textbf{W}^{(t)\top}+\beta\vec{\textbf{X}}^{(t)}\textbf{U}^{(t)\top}+ \theta\vec{\textbf{Z}}^{(t)}\textbf{V}^{(t)\top}+\mu\vec{\textbf{X}}^{(t)}\textbf{P}^{(t)}$.
Since $\vec{\textbf{B}}^{(t)}$ is discrete, it is difficult to optimize. To solve this problem, we adopt the discrete cyclic coordinate descent algorithm \cite{shen2015supervised} to learn $\vec{\textbf{B}}^{(t)}$ bit-by-bit iteratively until convergence. Due to the page limit, we directly give the solution,
\begin{equation} \label{Y1-solution}
\begin{split}
&\vec{\textbf{b}}^{(t)\top}=sign\Big(\textbf{q}^{(t)\top}-\textbf{w}^{(t)}\textbf{W}^{(t)'\top}\vec{\textbf{B}}^{(t)'\top}\textbf{K}^{(t)}\\
&-\beta\textbf{u}^{(t)}\textbf{U}^{(t)'\top}\vec{\textbf{B}}^{(t)'\top}-\theta\textbf{v}^{(t)}\textbf{V}^{(t)'\top}\vec{\textbf{B}}^{(t)'\top}\big),
\end{split}
\end{equation}
where $\vec{\textbf{b}}^{(t)\top}$ denotes the $l$-th row of $\vec{\textbf{B}}^{(t)}$, $l \in\{ 1, 2, {...}, r\}$, $\vec{\textbf{B}}^{(t)'}$ is the submatrix of $\vec{\textbf{B}}^{(t)}$ excluding $\vec{\textbf{b}}^{(t)\top}$, similarly, $\textbf{q}^{(t)\top}$ denotes the $l$-th row of $\textbf{Q}^{(t)}$; $\textbf{w}^{(t)\top}$ denotes the $l$-th row of $\textbf{W}^{(t)}$, $\textbf{W}^{(t)'}$ is the submatrix of $\textbf{W}^{(t)}$ excluding $\textbf{w}^{(t)\top}$; $\textbf{u}^{(t)\top}$ denotes the $l$-th row of $\textbf{U}^{(t)}$, $\textbf{U}^{(t)'}$ is the submatrix of $\textbf{U}^{(t)}$ excluding $\textbf{u}^{(t)\top}$; $\textbf{v}^{(t)\top}$ denotes the $l$-th row of $\textbf{V}^{(t)}$, $\textbf{V}^{(t)'}$ is the submatrix of $\textbf{V}^{(t)}$ excluding $\textbf{v}^{(t)\top}$. At the $t$-th round, each bit $\vec{\textbf{b}}^{(t)}$ of newly coming data is iteratively updated based on the pre-learnt matrix $\vec{\textbf{B}}^{(t)'}$ until the procedure converges to a set of better codes $\vec{\textbf{B}}^{(t)}$.

\subsection{Analysis}\label{time}
At the $t$-th round, the time complexity for updating $\textbf{W}^{(t)}$, $\textbf{U}^{(t)}$, $\textbf{V}^{(t)}$, and $\textbf{P}^{(t)}$ is $\mathcal{O}((n_tr^2+n_tcr+r^3)T)$, $\mathcal{O}((n_tr^2+r^3+n_trd+r^2d)T)$, $\mathcal{O}((n_tr^2+r^3+n_trf+r^2f)T)$, and $\mathcal{O}((n_td^2+d^3+n_trd+d^2r)T)$, respectively. The complexity for obtaining $\vec{\textbf{B}}^{(t)}$ is $\mathcal{O}((cr^2+dr^2+fr^2+n_tr^2)gT)$. Thereinto, $n_t$ is the size of the newly coming data, $r$ is the hash code length, $g$ is the iteration of DCC, $c$ is the amount of tags, $f$ is the dimensionality of word2vec embedding features, $d$ is the dimensionality of visual features, and $T$ is the number of iterations. Therefore, the overall computational complexity is linear to the size of the newly coming data $n_t$, which makes our method scalable.
The whole optimization algorithm is summarized in Algorithm \ref{algorithm}.
\begin{algorithm}[tb]
  \caption{Online optimization of WOH at round $t$.}
  \label{algorithm}
  \begin{algorithmic}
    \State
      $\textbf{Input: }$ new data chunk: $\phi{(\vec{\textbf{X}}^{(t)})}$ and $\vec{\textbf{Y}}^{(t)}$; accumulated data (old data): $\phi{(\tilde{\textbf{X}}^{(t)})}$, $\tilde{\textbf{Y}}^{(t)}$, $\tilde{\textbf{Z}}^{(t)}$, and $\textbf{W}^{(t-1)}$.
    \State
      $\textbf{Output: }$ Hash codes and hash function.
    \State \textbf{Procedure:}
    \State Generate the image level semantic representation $\vec{\textbf{Z}}^{(t)}$.
    \State Randomly initialize $\vec{\textbf{B}}^{(t)}$ and ${\textbf{W}}^{(t)}$.
    \For{$t=1,{...}, T$}

            \State Update $\textbf{U}^{(t)}$ according to Eq.(\ref{U_2});
            \State Update $\textbf{P}^{(t)}$ according to Eq.(\ref{P_2});
            \State Update $\textbf{V}^{(t)}$ according to Eq.(\ref{V_2});
            \State Update $\textbf{W}^{(t)}$ according to Eq.(\ref{W4});
            \State Update $\vec{\textbf{B}}^{(t)}$ iteratively according to Eq.(\ref{Y1-solution});
    \EndFor
  \end{algorithmic}
\end{algorithm}

\section{Experiments}

\subsection{Datasets and Evaluation metric}
\textbf{MIRFlickr} \cite{huiskes2008mir} consists of $25,000$ images collected from Flickr associated with $1,386$ user-provided tags. The tags appearing less than $50$ times are removed. We further removed those tags which cannot be transformed into the embedding space, such as ``2007", ``i500", and ``d200"; and those images with no tags are omitted. Finally, $17,833$ images are left. We randomly split the dataset into a query set with $1,000$ images and the remaining are served as the training set. To support the online learning, the training set further divided into $8$ data chunks with each of the first $7$ chunks containing $2,000$ samples and the last chunk containing $2,833$ instances.

\textbf{NUS-WIDE} \cite{chua2009nus} contains $269,648$ images collected from Flickr by the Lab for Media Search in the National University of Singapore. This dataset is large-scale and covers $5,018$ unique tags. $194,541$ social images, which correspond to the $21$ most frequent labels, are left. Following \cite{lin2020hadamard}, we randomly split the data set into a query set with $2,000$ images and a retrieval set with the others, and further randomly picked out $40,000$ samples from retrieval set as the training set. In order to support the online learning, we split the training set into $8$ chunks and per chunk contains $5,000$ instances.

For both datasets, the $4,096$-D output of the pre-trained VGG-F, which is trained on the ImageNet dataset, is used to represent images. If two images share at least one ground-truth label, they are similar; otherwise, they are semantically dissimilar. Note that, during training, only user-generated tags and image features are used while the ground-truth labels are only leveraged during evaluation.

We employed the widely-used criteria, i.e., mean average precision (MAP) to evaluate the retrieval performance. The larger value indicates the better retrieval performance.

\subsection{Baselines and Implementation details}
Seven state-of-the-art hashing models are selected for comparison: 1) traditional methods, i.e., SH \cite{weiss2009spectral}, SDH \cite{shen2015supervised}, and COSDISH \cite{kang2016column}; 2) social image hashing, i.e., WDH \cite{cui2020efficient}; 3) online hashing, i.e., OSH \cite{leng2015online}, BOSDH \cite{lin2019towards}, and HMOH \cite{lin2020hadamard}.

Except for OSH, BOSDH, and HMOH, other baselines are batch-based and their hash functions and hash codes are
retrained on all accumulated data at each round. Compared to online methods, the training of traditional deep hashing is not practical for online retrieval \cite{wu2019deep}. Thus, no deep weakly-supervised hashing methods are adopted.


We set the parameter $\alpha$, $\beta$, $\theta$, $\mu$, and $T$ to $300$, $0.1$, $0.1$, $10$, and $7$ according to parameter experiments. Moreover, the amount of anchor points $m$ is $1,000$ and the dimensionality of embedding feature denoted as $f$ is $300$, and the iterations of DCC g is $3$.

\begin{figure}[t]
\begin{minipage}{0.49\linewidth}\centering
\centerline{\includegraphics[height=3.9cm]{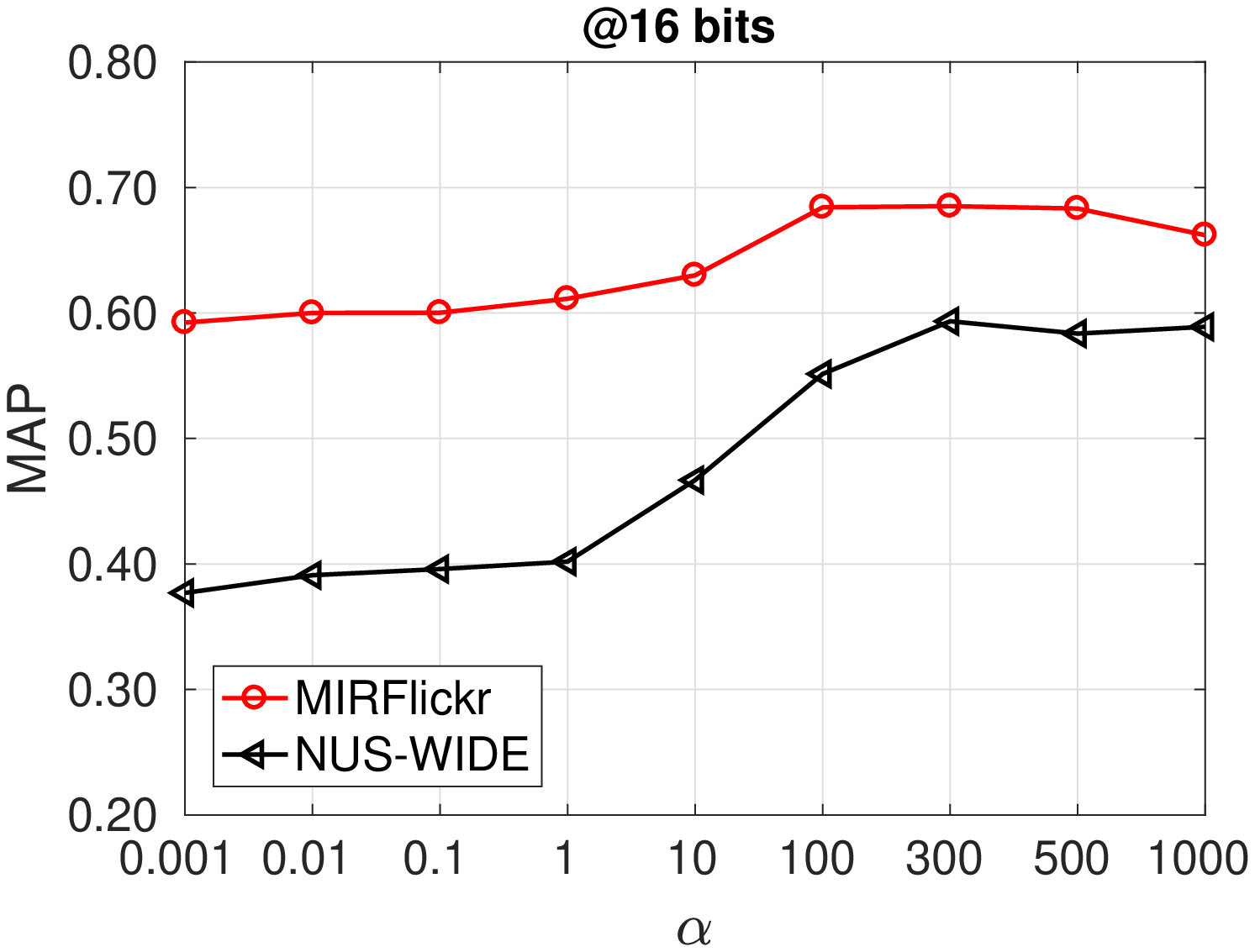}}
\end{minipage}
\begin{minipage}{0.49\linewidth}\centering
\centerline{\includegraphics[height=3.9cm]{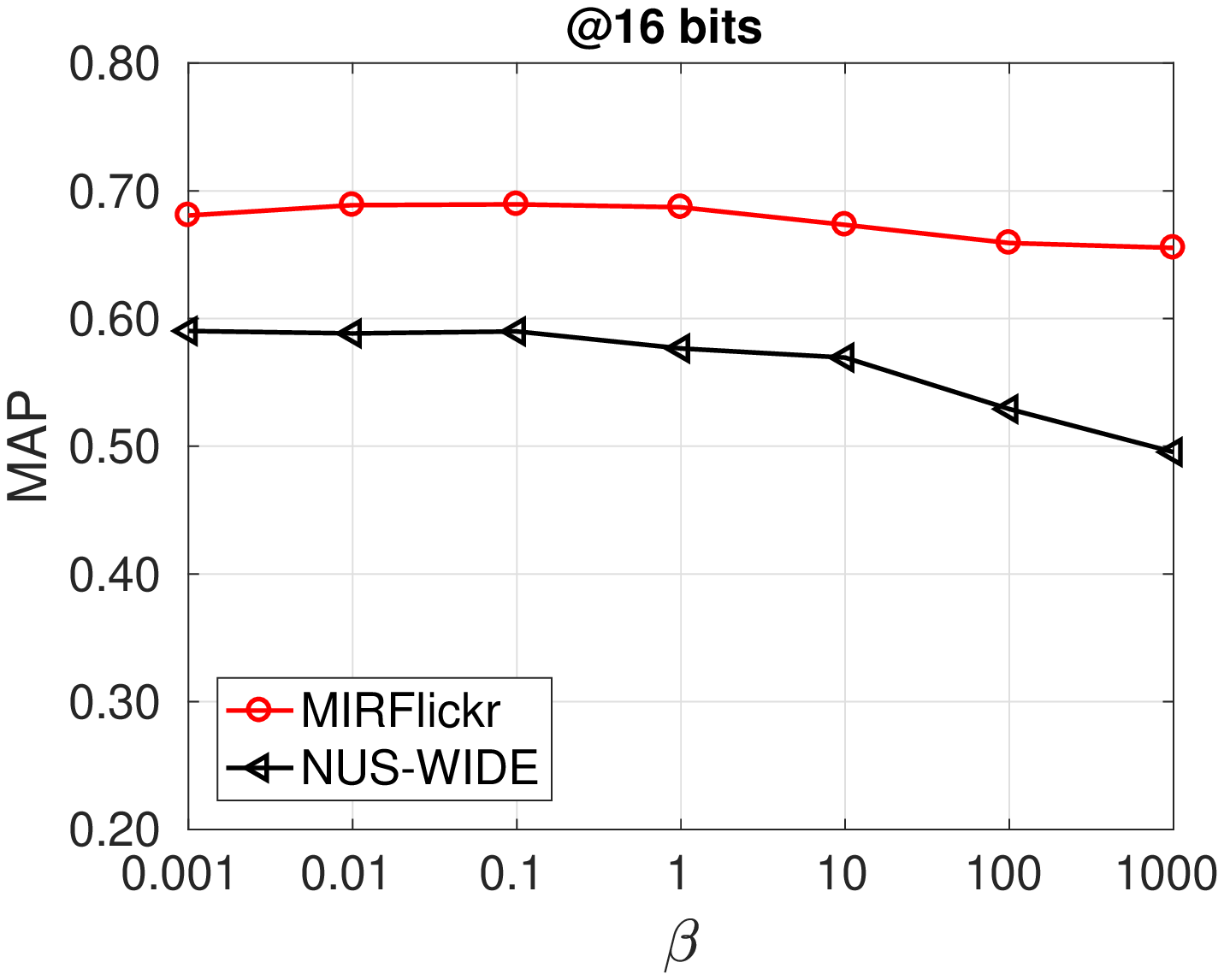}}
\end{minipage}
\begin{minipage}{0.49\linewidth}\centering
\centerline{\includegraphics[height=3.9cm]{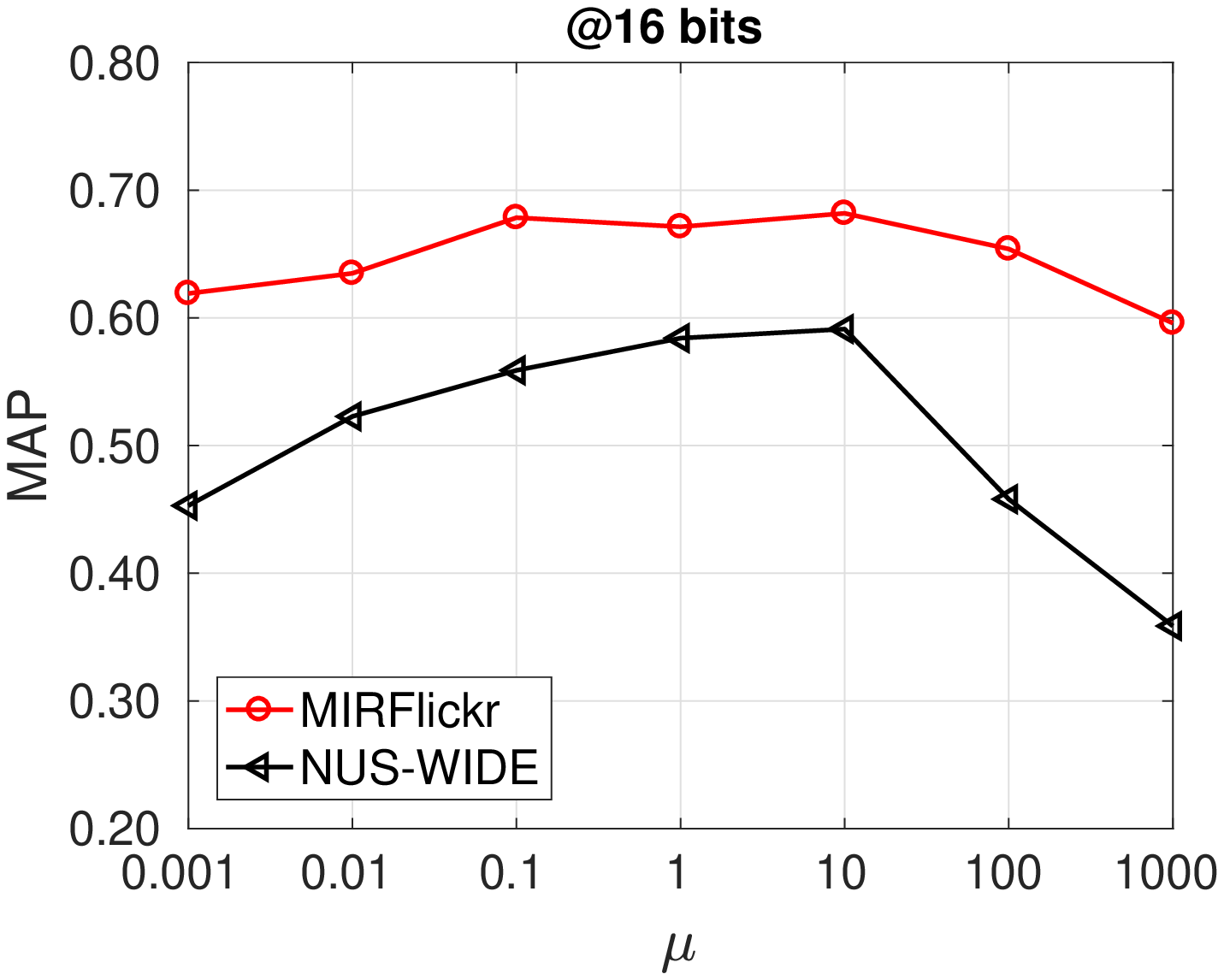}}
\end{minipage}
\begin{minipage}{0.49\linewidth}\centering
\centerline{\includegraphics[height=3.9cm]{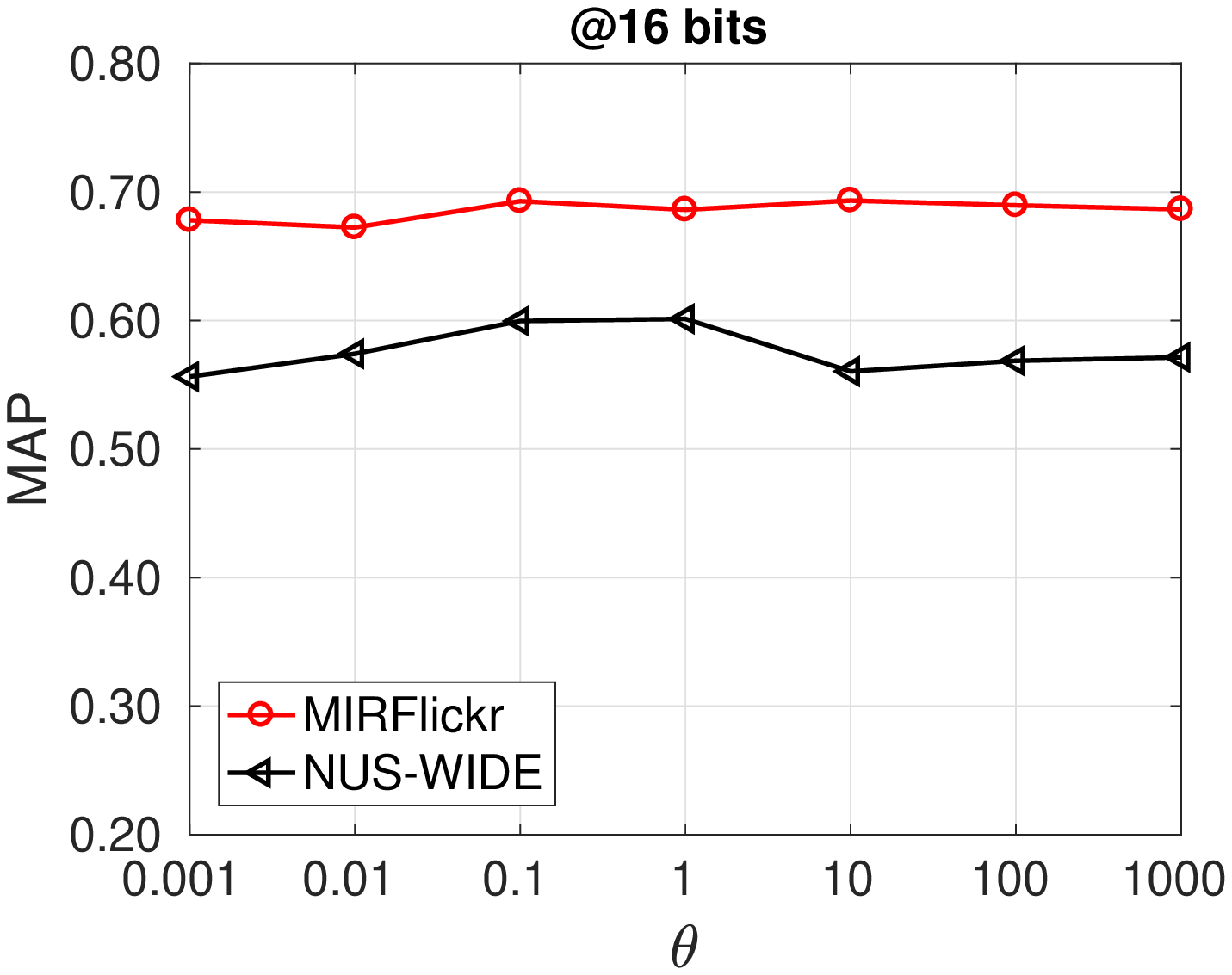}}
\end{minipage}
\caption{Sensitivity analysis of parameters $\alpha$, $\beta$, $\mu$, and $\theta$.}\label{fig-param}\medskip\end{figure}

\subsection{Comparison with Baselines}
Table \ref{map-table} lists the MAP results of WOH and all the comparison methods at the last rounds. The MAP results of all methods at each round with 16 bits and 64 bits are plotted in Fig.\ref{MAP-round}. From these results, we can observe that:

1) Although the superiority of supervised hashing methods over unsupervised ones is found by many literatures, such phenomenon is not obvious in the weakly-supervised case. We can easily find that some unsupervised methods can offer better accuracy than the supervised ones, e.g., OSH outperforms BOSDH and HMOH. And the reason may be that the weakly-supervised information, i.e., tags, is weak and noisy.

2) Our method outperforms WDH, which is specifically designed for weakly-supervised social image retrieval task. Besides, WDH also uses the $\ell_{2,1}$-norm. Compared to WDH, we can find the performance of WOH is better, further confirming that our WOH can better harness the tags and learn better hash codes in the online manner.

3) Compared with all online hashing methods, i.e., OSH, BOSDH, and HMOH, our method always offers better performance. This phenomenon also shows that our model can effectively remove the noise in tags and benefit from the weak supervision information.

4) WOH outperforms all the adopted state-of-the-art baselines, demonstrating its effectiveness.

In summary, our WOH works well for retrieving weakly-supervised social images in an online manner.

\begin{figure}[t]
\begin{minipage}{0.57\linewidth}\centering
\centerline{\includegraphics[height=3.2cm]{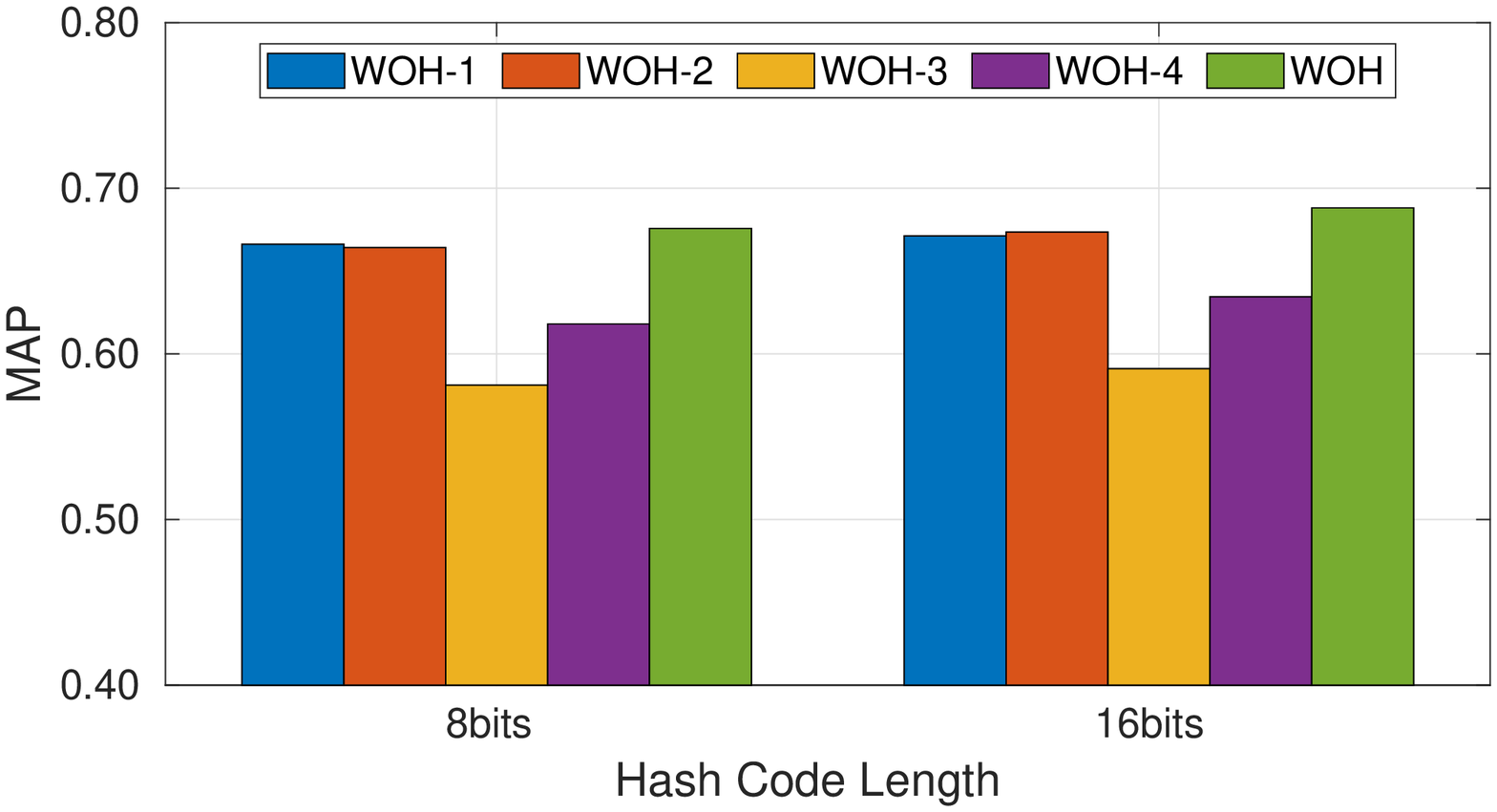}}
{(a) ablation}
\end{minipage}
\begin{minipage}{0.42\linewidth}\centering
\centerline{\includegraphics[height=3.2cm]{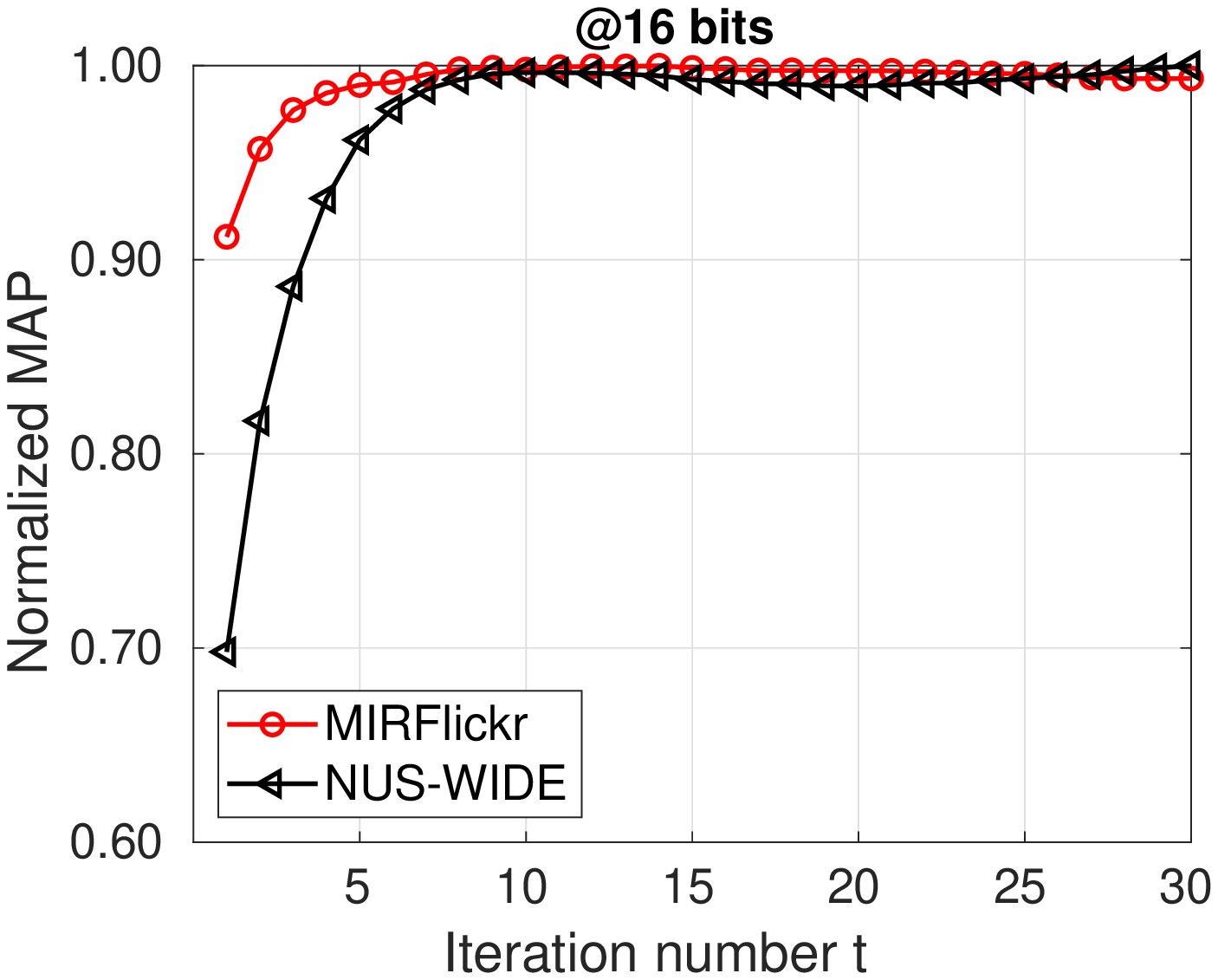}}
{(b) convergence}
\end{minipage}
\caption{The results of ablation and convergence experiments.}
\label{mix-fig}
\end{figure}

\begin{table}[tb]\scriptsize \setlength{\belowcaptionskip}{-1mm}  
\caption{Training time (seconds) on MIRFlickr.}
\label{time-online-table}
\centering\begin{tabular}{c ccccc }\toprule[0.8pt]
{Method}& chunk1 & chunk2 & chunk3 & chunk4 & chunk5 \\\hline
{OSH} &{1.0173}&{1.0388}&{1.0391}&{1.0319}&{1.0100}\\
{BOSDH}&{0.0665}&{12.0278}&{22.9630}&{34.0123}&{51.3250}\\
{HMOH}&{0.2913}&{0.2509}&{0.2560}&{0.2640}&{0.2712}\\
{WOH}&{1.7246}&{1.7026}&{1.7425}&{1.7382}&{1.7619}\\
\toprule[0.8pt]\end{tabular}\end{table}

\subsection{Further Analysis}
\textbf{{Parameter Sensitivity Analysis:}}
To analyze the influence of parameters on the performance, we conducted experiments on MIRFlickr and NUS-WIDE in the case of 16-bit code length and the results are plotted in Fig.\ref{fig-param}. It can be seen that the search performance is strongly related to the parameters $\alpha$ and $\mu$. The performance maintains satisfactory when $\alpha$ ranges from $300$ to $500$ and $\mu$ ranges from $1$ to $10$ on both two datasets. It also can be found that WOH is robust to parameters $\beta$ and $\theta$. Thus, we set $\alpha$, $\beta$, $\mu$, and $\theta$ to $300$, $0.1$, $10$, and $0.1$, respectively.

\textbf{Ablation Experiments:} To provide ablation analysis, four derivatives of our model are designed and the experimental results on MIRFlickr is presented in Fig.\ref{mix-fig}-(a). WOH-1 denotes the variant that sets $\theta=0$; WOH-2 sets $\theta=0$ and omits $\mathcal{O}_{reg}$ in Eq. $(\ref{overall})$; WOH-3 represents the variant which sets $\alpha$ to $0$; for the last derivative WOH-4, both $\beta$ and $\mu$ are set to $0$ and the hash functions are learnt by a two-step hashing strategy \cite{Lin2013AGT}. From this figure, we can find: 1) WOH outperforms WOH-1 and WOH-2, demonstrating that by elaborately learning from weakly-supervision, i.e., user-provided tags, better hash codes can be obtained; 2) WOH outperforms WOH-3, showing the necessity of regularization; 3) WOH-4 learns only from tags and performs worse than WOH, revealing the importance of visual information.

\textbf{{Convergence:}}
We validated the convergence of the proposed alternative optimization algorithm by experiments. Fig.\ref{mix-fig}-(b) illustrates the convergence curves of WOH based on the first data chunk in the case of 16-bit on MIRFlickr and NUS-WIDE. From this figure, we can see that WOH converges quickly. Considering both efficiency and performance, we chose $T=7$ in all experiments.

\textbf{{Time Analysis:}} As shown in Section \ref{time}, the time complexity of WOH is linearly dependent on the size of newly coming data $n_t$. To quantitatively evaluate the efficiency of WOH, we further conducted experiments and the time cost of four online hashing methods on MIRFlickr with the 16-bit codes is listed in Table \ref{time-online-table}. We can observe that: 1) The unsupervised OSH holds the best training efficiency because it leaves tags out of consideration while supervised and weakly-supervised methods need to spend extra time handling tags. 2) Although the reported training time of HMOH is little, it takes a lot of time to calculate the Hadamard Matrix, which is not included in the training time. 3) Considering both efficiency and effectiveness, WOH is the best choice.


\section{Conclusion}\label{section-conclusion}
In this paper, we present a novel hashing method named Weakly-supervised Online Hashing, which is specially designed for retrieving weakly labeled social images in online fashion. To the best our knowledge, it is the first attempt to apply the idea of online hashing to weakly-supervised social image retrieval. To learn more accurate hash codes, WOH explores the weak supervision by considering the semantics of tags and removing the noise. Extensive experiments on two real-world social image datasets have been conducted and the results demonstrate the superiority of WOH over the state-of-the-art baselines.

\bibliographystyle{IEEEbib}
\bibliography{icme2021template}

\end{document}